\documentclass[runningheads]{llncs}
 
\usepackage{eccv}

\usepackage{eccvabbrv}

\usepackage{graphicx}
\usepackage{booktabs}

\usepackage{multirow}

\usepackage{hyperref}

\usepackage{orcidlink}

\begin{document}

\title{Partial Skeleton Visibility for Action Recognition: A Constrained Field-of-View Approach} 

\titlerunning{Partial Skeleton Visibility for Action Recognition}

\author{Yingjie Dai\inst{1}\orcidlink{0009-0009-3631-5024} \and
Tianyang Xu\inst{1}\thanks{Corresponding author.}\orcidlink{0000-0002-9015-3128} \and
Yanglin Deng\inst{1}\orcidlink{0009-0005-7022-2869} \and \\
Xiao-Jun Wu\inst{1}\orcidlink{0000-0002-0310-5778} \and
Josef Kittler\inst{2}\orcidlink{0000-0002-8110-9205}}

\authorrunning{Y. Dai et al.}

\institute{Jiangnan University, Wuxi, China \and
University of Surrey, Guildford, UK\\
\email{daiyingjie2024@gmail.com;
\{tianyang.xu,wu\_xiaojun\}@jiangnan.edu.cn;
yanglin\_deng@163.com;
j.kittler@surrey.ac.uk;}
}

\maketitle

\begin{abstract}

Skeleton-based action recognition has achieved remarkable success by exploiting joint coordinates and their topological connections, yet prevailing methods overwhelmingly assume complete and clean skeleton inputs. 
In real-world deployments, such as egocentric vision, crowded surveillance, wearable devices, or edge robotics, limited field-of-view (FoV) frequently causes substantial joint visibility dropout, leading to severe performance degradation that existing models are largely unprepared to handle.
To bridge this critical yet underexplored gap, we introduce PartialVisGraph, a novel hypergraph framework tailored for robust skeleton action recognition under constrained FoV. 
We first construct highly expressive hypergraphs by introducing learnable virtual hyperedges that form a soft incidence matrix, capturing flexible high-order dependencies beyond conventional pairwise graphs. 
We then propose the Single-Head Sample-Adaptive Transformer, which adaptively aggregates joint features onto hyperedges while explicitly incorporating a visibility prior. 
This prior selectively gates information flow, preventing occluded or out-of-view joints from corrupting reliable feature propagation. 
We further establish rigorous evaluation protocols with realistic FoV simulation benchmarks on NTU RGB+D 60 and 120.
Extensive experiments demonstrate that PartialVisGraph consistently achieves state-of-the-art accuracy under partial visibility, with gains of up to 68.8\% on subsets with severe FoV restrictions compared to recent strong baselines, while remaining superior on full-visibility settings. 
Our approach offers a principled and practical pathway toward deployable skeleton-based action understanding in unconstrained environments. Resources related to the constrained FoV setting used in this work are available at:
\url{https://github.com/yaa1haa1/PartialVisGraph}.
\end{abstract}

\section{Introduction}
\label{sec:intro}

\begin{figure}[tb]
  \centering
  \includegraphics[width=\linewidth]{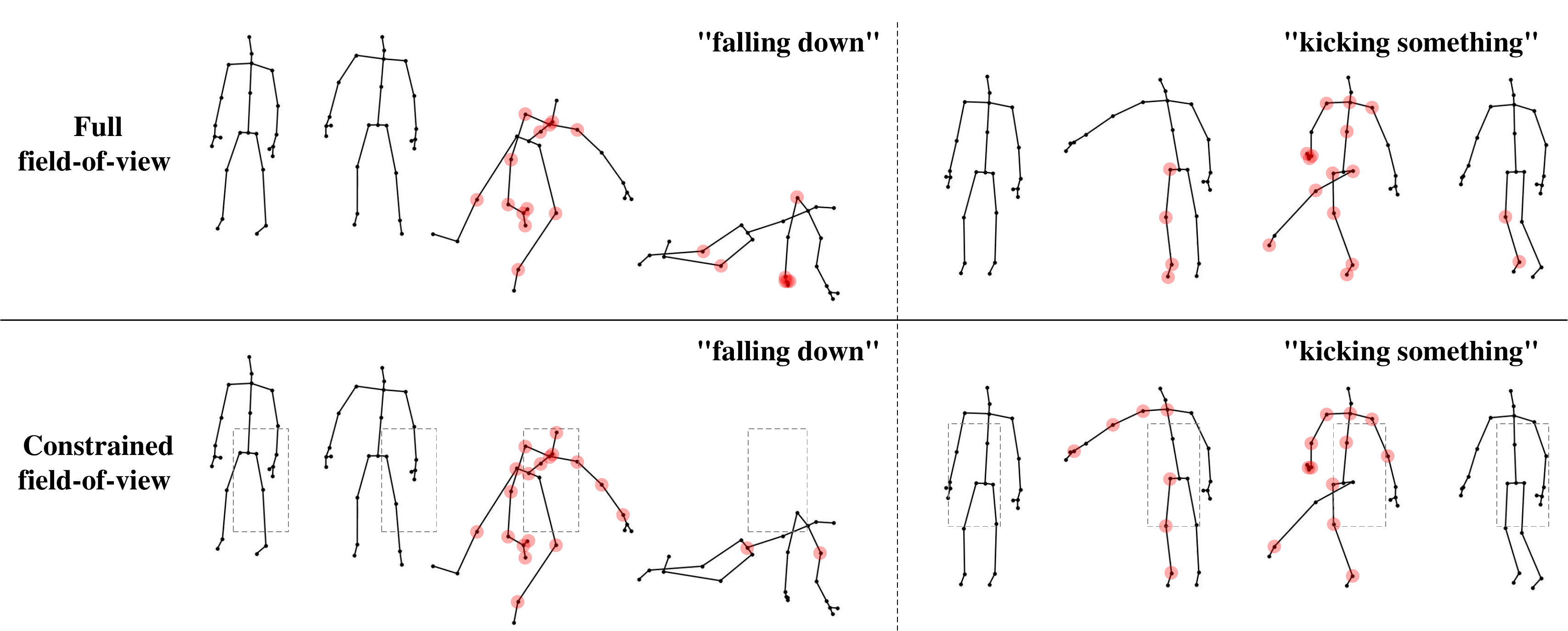}
  \caption{Skeleton-based action recognition under constrained field-of-view. The dashed box indicates the restricted visible region, and joints whose features in the final layer contribute most to the model’s prediction are highlighted in red.}
  \label{fig1}
\end{figure}

Recently, skeleton-based human action recognition \cite{presti20163d,ren2024survey,sun2022human,zhu2025semantic,zheng2024towards,chen2025neuron,zhu2024part} has demonstrated significant potential across a wide range of applications, including medical rehabilitation, virtual reality, intelligent surveillance, and autonomous driving. By representing human motion through joints and their topological structure, skeleton sequences provide a structured and geometry-aware description of body dynamics. This representation directly captures the kinematic relationships among joints and the temporal evolution of movements. Owing to its abstraction from appearance, skeleton data is inherently more robust to complex backgrounds, illumination variations, and viewpoint changes. Meanwhile, its compact and low-dimensional structure substantially reduces computational overhead. With the continuous advancement of depth sensors and 3D pose estimation algorithms \cite{martinez2017simple,pavllo20193d,banik2023occlusion,choi2020pose2mesh,hao2025perspose}, the accuracy and stability of skeleton acquisition have steadily improved, further accelerating the development of skeleton-based action recognition in both theoretical research and real-world deployment.

Looking ahead, within the emerging paradigm of embodied intelligence \cite{sun2024comprehensive}, reliable human action recognition is a prerequisite for natural human-machine interaction \cite{mehak2024action}, for example, in service robotics and assistive devices. In embodied settings, sensors are often mounted on moving agents or constrained platforms, so camera observations are commonly partial and occluded. Methods \cite{do2024skateformer,yan2024crossglg,li2024sa,ma2022learning,li2024skeleton,do2025bridging,song2020richly} that assume complete skeletons are therefore ill-suited for real-world embodied deployment, motivating our work to explicitly address the constrained FoV setting.

From a methodological perspective, the human skeleton naturally forms a standard graph \cite{yan2018spatial}, where joints and their physical connections correspond to vertices and edges, respectively. Most existing approaches \cite{chi2022infogcn,cheng2020decoupling,duan2022pyskl,lee2023hierarchically,gao2025skeleton,wang2022skeleton,chen2021multi,yu2023shuffle} are therefore designed upon this normal graph structure. They typically employ graph convolutions to exchange information among spatially connected joints within a single frame, and temporal convolutions to capture motion dynamics across frames. However, in a normal graph, each edge connects only two vertices, and information propagates primarily along physical connections \cite{shi2019two}. As a result, modelling relationships between physically distant joints (\eg, the two hands during clapping) can be suboptimal. In reality, human actions are jointly defined by multiple body parts \cite{thakkar2018part}. Beyond pairwise interactions, many actions involve higher-order relationships among several joints (\eg, standing up requires coordinated movement of the legs, spine, and other joints). Under constrained FoV conditions, it becomes even more critical to reason about information flow from a multi-joint perspective. Joint modelling over multiple joints may enable the aggregation of weak and spatially distributed cues into discriminative composite features.

To address these challenges, we adopt hypergraphs for skeleton-based action recognition under constrained FoV. Only a few prior works \cite{zhou2025adaptive,zhou2022hypergraph} have explored hypergraphs for this task, and these existing methods often have limitations in hypergraph construction. To this end, we propose a novel hypergraph construction strategy and introduce a Single-Head Sample-Adaptive Transformer (SHSAT) to aggregate node features onto hyperedges. In addition, we explicitly constrain information propagation to better suit constrained FoV scenarios. 

As illustrated in \cref{fig1}, our method continues to extract useful discriminative information even under a constrained FoV. For example, in the "kicking something" sample, the model perceives the right arm lifting and moving out of view, a motion used to maintain balance during a kick, and it is able to extract additional information from the otherwise weak right-arm joint signals. 
In summary, our contributions are as follows:
\begin{itemize}
    \item We are the first to introduce the task of skeleton-based action recognition under constrained FoV, which relaxes the deployment constraints of action recognition systems and provides a practical solution for potential field-of-view limitations in future embodied intelligence scenarios.
    
    \item We propose a novel hypergraph construction scheme and design a corresponding feature aggregation method tailored to this formulation. 
    
    \item Our model achieves superior performance compared with state-of-the-art methods under constrained FoV settings. In addition, when trained and evaluated under full FoV conditions, it also surpasses existing approaches.
\end{itemize}

\section{Related Works}

\subsection{GCN-Based Approaches}
Since the physical topology of human joints and bones can be naturally represented as a graph, Graph Convolutional Networks (GCNs) \cite{kipf2016semi} have been widely adopted for skeleton-based action recognition. \cite{yan2018spatial} was the first to introduce GCNs to jointly model spatial and temporal features, demonstrating their strong performance for action recognition. Subsequent studies observed that a fixed topology defined by natural anatomical connections is inherently limited. As a result, most later works \cite{chen2021channel,lee2023hierarchically,zhou2024blockgcn,liu2025revealing,ye2020dynamic,chan2020gas} focused on learning adaptive graph structures. \cite{shi2019two} incorporated adaptive graph convolution to dynamically learn the underlying topology, achieving superior performance compared to earlier methods.

However, these approaches typically model relationships between joints as strictly pairwise interactions. In reality, human actions arise from coordinated movements of multiple joints, which entail not only binary dependencies between joint pairs but also complex higher-order interactions among groups of joints.

\subsection{Hypergraph-Based Approaches}
Recognising that pairwise connections are insufficient to capture collaborative interactions among multiple joints, recent studies have introduced hypergraphs \cite{feng2019hypergraph,bai2021hypergraph} into skeleton-based action recognition. \cite{zhou2022hypergraph} constructs a hypergraph and proposes a hypergraph-based self-attention mechanism (HyperSA) to explicitly model high-order dependencies among joints, thereby overcoming the limitation of normal graph convolution, which is restricted to pairwise modelling. Similarly, \cite{zhou2025adaptive} adaptively optimises the hypergraph structure during training, enabling joint-level feature aggregation and relational modelling at a higher semantic level, and dynamically uncovering action-driven high-order correlations.

Despite their promising performance, these methods primarily target action recognition with complete skeleton data, overlooking skeleton inputs under constrained FoV conditions that commonly arise in real-world scenarios and future embodied intelligence applications. Moreover, their hypergraph construction comes with structural constraints: \cite{zhou2022hypergraph} restricts each node to belong to only one hyperedge, while \cite{zhou2025adaptive} requires the number of hyperedges to equal the number of joints.

\subsection{Occluded Human Action Recognition}
Occluded skeleton-based human action recognition has attracted increasing attention in recent years. \cite{shi2023occlusion} designs a multi-branch architecture tailored to different occlusion scenarios and integrates multi-scale motion features, effectively improving recognition performance under occluded skeleton conditions. \cite{chen2023occluded} addresses performance degradation caused by occlusion or noise by introducing a dual inhibition training strategy. Through simulating key parts and random occlusions, combined with global-local part modelling, the method significantly enhances robustness to incomplete skeleton inputs.

Although these approaches improve robustness to occlusion to a certain extent, they still exhibit limitations in terms of architectural complexity and generalisation capability. More importantly, their focus is on occlusion-specific scenarios rather than skeleton data under constrained FoV settings, which fundamentally distinguishes them from our work.

\section{Method}
\subsection{Preliminaries}
\subsubsection{Graph Convolutional Network.}
In skeleton-based action recognition, the human body is naturally represented as a graph, where joints are treated as nodes and bones as edges. 
Let $\mathbf{A} \in \mathbb{R}^{N \times N}$ denote the adjacency matrix of the skeleton graph with $N$ joints. Following the standard formulation, we first add self-connections to obtain $\tilde{\mathbf{A}} = \mathbf{A} + \mathbf{I}$, where $\mathbf{I}$ is the identity matrix. The corresponding degree matrix is defined as $\tilde{\mathbf{D}} \in \mathbb{R}^{N \times N}$ with diagonal entries
\(
\tilde{D}_{ii} = \sum_{j} \tilde{A}_{ij}.
\)
Given the input feature matrix $\mathbf{X} \in \mathbb{R}^{N \times C_{\text{in}}}$, a graph convolutional layer can be written as
\begin{equation}
\mathbf{F} = \sigma \left( \tilde{\mathbf{D}}^{-\frac{1}{2}} 
\tilde{\mathbf{A}} 
\tilde{\mathbf{D}}^{-\frac{1}{2}} 
\mathbf{X} 
\mathbf{W} \right),
\end{equation}
where $\mathbf{W} \in \mathbb{R}^{C_{\text{in}} \times C_{\text{out}}}$ is a learnable weight matrix for feature transformation, and $\sigma(\cdot)$ denotes a nonlinear activation function.

\subsubsection{Hypergraph.}
A hypergraph is a generalised form of a graph that allows each edge to connect an arbitrary number of vertices. Formally, a hypergraph is defined as $\mathcal{G}_h = (\mathcal{V}, \mathcal{E})$, where $\mathcal{V}$ is a finite set of vertices and $\mathcal{E}$ is a set of non-empty subsets of $\mathcal{V}$, referred to as hyperedges. Unlike normal graphs, where each edge connects exactly two vertices, a hyperedge can connect any number of vertices, enabling the modelling of higher-order relationships.

Analogous to the adjacency matrix in a standard graph, a hypergraph is commonly represented by an incidence matrix $\mathbf{H} \in \mathbb{R}^{N \times K}$, where $N = |\mathcal{V}|$ denotes the number of vertices and $K = |\mathcal{E}|$ denotes the number of hyperedges. The entries of the incidence matrix are defined as
\begin{equation}
\mathbf{H}_{v,e} =
\begin{cases}
1, & \text{if } v \in e, \\
0, & \text{otherwise}.
\end{cases}
\end{equation}

\begin{figure}[tb]
  \centering
  \includegraphics[width=\linewidth]{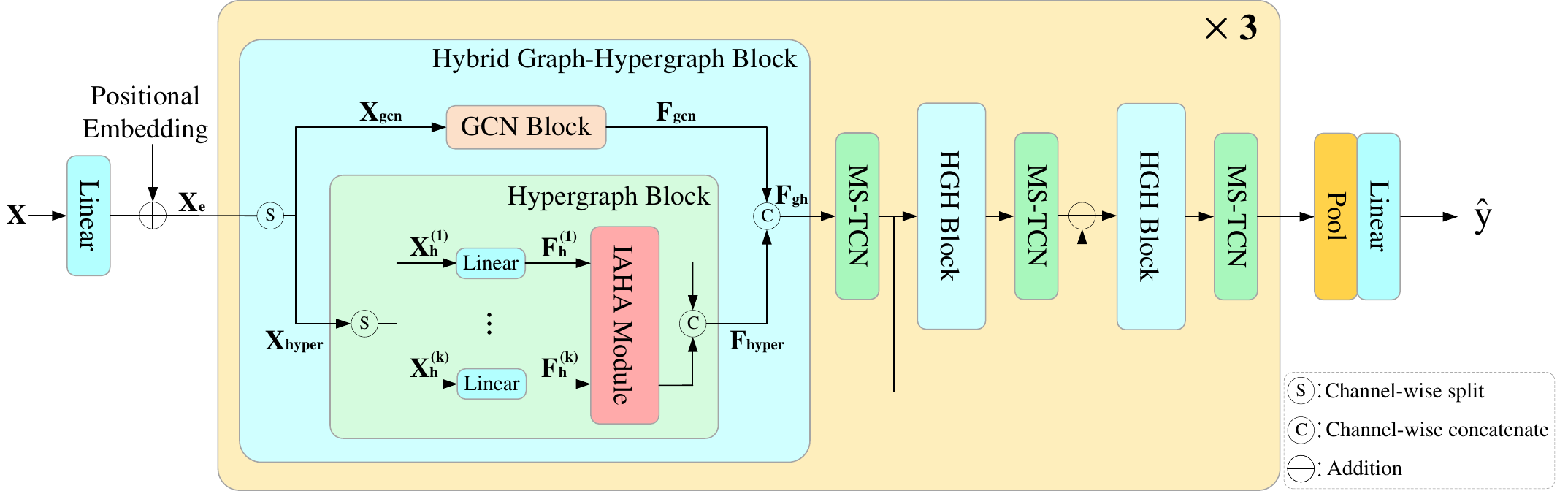}
  \caption{Overview of our PartialVisGraph approach}
  \label{fig2}
\end{figure}

\subsection{Overview}
The overall architecture of our method is illustrated in \cref{fig2}. 
Given a skeleton sequence $\mathbf{X} \in \mathbb{R}^{T,V,C_r}$ as input, where $T$ denotes the number of frames, $V$ the number of joints per frame, and $C_r$ the number of input channels, we first project the low-dimensional raw skeleton data into a higher-dimensional feature space through a linear layer. The positional embedding is then added to the projected features to preserve structural and temporal information, resulting in the embedded representation $\mathbf{X}_e \in \mathbb{R}^{T,V,C_f}$, where $C_f$ denotes the feature dimension after projection.
The embedded features are subsequently fed into a Hybrid Graph-Hypergraph block (HGH block), which models spatial relationships among joints within each frame. To capture temporal dependencies across multiple frames, we employ a Multi-Scale Temporal Convolutional Network (MS-TCN) \cite{liu2020disentangling}. After stacking multiple HGH blocks and MS-TCN layers, the final feature representation is subjected to temporal pooling followed by a linear projection to produce the prediction output $\hat{\mathbf{y}} \in \mathbb{R}^{N_{\text{class}}}$, where $N_{\text{class}}$ denotes the number of action categories.

\subsection{Hybrid Graph-Hypergraph Block}
\begin{figure}[tb]
  \centering
  \begin{subfigure}{0.48\linewidth}
    \centering
    \includegraphics[width=\linewidth]{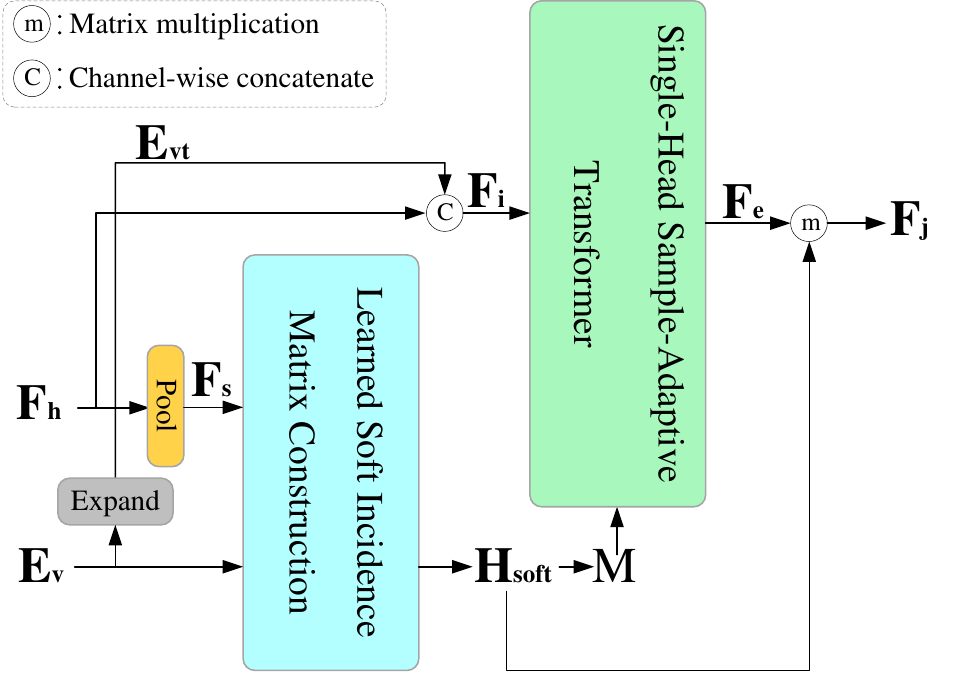}
    \caption{Internal architecture of the IAHA module}
    \label{fig3a}
  \end{subfigure}
  \hfill
  \begin{subfigure}{0.48\linewidth}
    \centering
    \includegraphics[width=\linewidth]{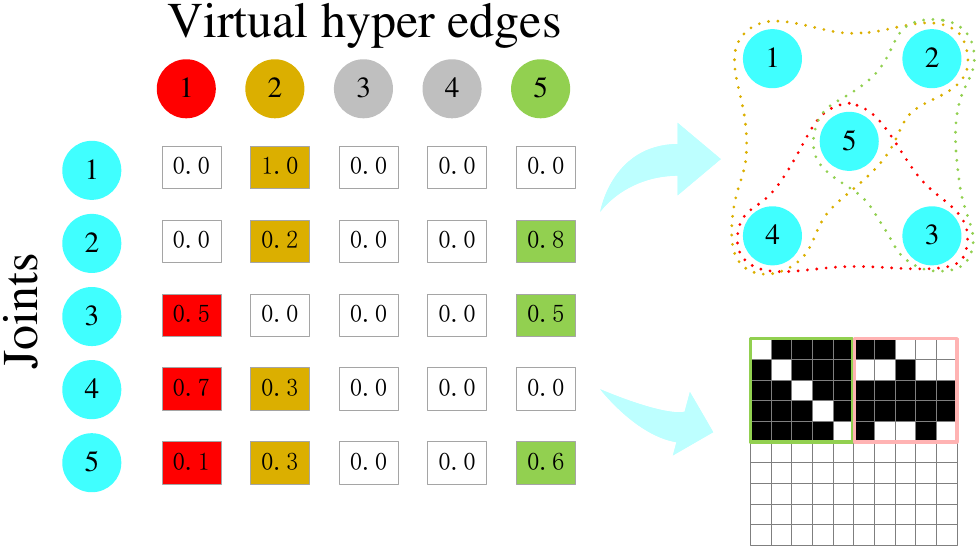}
    \caption{Construction of the soft incidence matrix and the corresponding attention mask}
    \label{fig3b}
  \end{subfigure}
  \caption{Incidence-aware hypergraph attention module}
  \label{fig3}
\end{figure}
To exploit the inherent topology of the human skeleton and thus provide a useful inductive bias, the HGH block splits the input $\mathbf{X}_e$ along the channel dimension into two parts, namely $\mathbf{X}_{\text{gcn}}$ with $C_f/4$ channels and $\mathbf{X}_{\text{hyper}}$ with $3C_f/4$ channels, and processes them with a GCN branch and a hypergraph branch respectively.

The GCN branch provides the explicit structural inductive bias. It treats the skeleton as a graph with adjacency $\mathbf{A}\in\mathbb{R}^{V\times V}$ and performs graph convolution, producing spatially-aware joint features $\mathbf{F}_{\text{gcn}}=\text{GCN}(\mathbf{X}_{\text{gcn}},\mathbf{A})$.

The hypergraph branch first evenly splits $\mathbf{X}_{\text{hyper}}$ into $k$ channel groups $\{\mathbf{X}_{\text{h}}^{(i)}\}_{i=1}^{k}$, each of shape $\mathbb{R}^{T\times V\times C_v}$ with $C_v=3C_f/(4k)$. Each group is linearly projected and fed to the Incidence-Aware Hypergraph Attention module (IAHA module), which is responsible for constructing the hypergraph in a data-driven manner, aggregating joint information onto hyperedges, and performing feature transformation via attention-style operations. The outputs of the IAHA module are concatenated to form $\mathbf{F}_{\text{hyper}}$.

Finally, we concatenate $\mathbf{F}_{\text{gcn}}$ and $\mathbf{F}_{\text{hyper}}$ along the channel dimension to obtain the block output $\mathbf{F}_{\text{gh}}\in\mathbb{R}^{T\times V\times C_{\text{gh}}}$, which is passed to subsequent modules.

\subsection{Incidence-Aware Hypergraph Attention Module}

As shown in \cref{fig3a}, the IAHA module introduces a set of learnable virtual hyperedge tokens $\mathbf{E}_v \in \mathbb{R}^{K_v \times C_v}$, where $K_v$ denotes the number of virtual hyperedges (we set $K_v=10$). Given an input feature $\mathbf{F}_h \in \mathbb{R}^{T\times V\times C_v}$, we first aggregate information along the temporal axis using average pooling to obtain a joint-level summary
\(
\mathbf{F}_s \;=\; \operatorname{Pool}_t(\mathbf{F}_h) \in \mathbb{R}^{V\times C_v},
\)
and then construct a soft incidence matrix $\mathbf{H}_{\mathrm{soft}}\in[0,1]^{V\times K_v}$ jointly from the joint summary $\mathbf{F}_s$ and the virtual hyperedge bank $\mathbf{E}_v$. Unlike a conventional incidence matrix whose entries are binary, the elements of $\mathbf{H}_{\mathrm{soft}}$ are continuous membership weights in the interval $[0,1]$. 
To limit computational cost, the same $\mathbf{H}_{\mathrm{soft}}$ is shared across all frames of a given sample.

Next, $\mathbf{H}_{\mathrm{soft}}$ is converted into an attention mask $\mathbf{M}\in\{0,1\}^{L\times L}$ with $L=K_v+V$, which encodes allowable attention links between the $K_v$ virtual hyperedge tokens and the $V$ joint tokens. This mask is applied during attention computation so that each joint token is allowed to transmit information only to the virtual hyperedge tokens it belongs to, ensuring that feature aggregation follows the learned hyperedge assignments.

To perform joint-hyperedge attention per frame, we replicate the virtual-token bank across the temporal dimension to form
\(
\mathbf{E}_{vt} \in \mathbb{R}^{T\times K_v \times C_v}
\) (\ie, $\mathbf{E}_{vt}$ contains $T$ identical copies of $\mathbf{E}_v$), and then concatenate the token sets to produce the attention input
\(
\mathbf{F}_i = \operatorname{concat}\big(\mathbf{E}_{vt},\; \mathbf{F}_h\big)
\in \mathbb{R}^{T\times (K_v+V)\times C_v}.
\)
A Single-Head Sample-Adaptive Transformer (SHSAT) is applied to $\mathbf{F}_i$ using the mask $\mathbf{M}$. From the attention output, we retain only the first $K_v$ tokens (the updated virtual hyperedge tokens) to obtain
\(
\mathbf{F}_e \in \mathbb{R}^{T\times K_v \times C_v}.
\)

Finally, the hyperedge features are redistributed to the joint domain using the soft incidence matrix. For each frame $t$, we compute
\(
\mathbf{F}_j^{(t)} \;=\; \mathbf{H}_{\mathrm{soft}}\,\mathbf{F}_e^{(t)} \in \mathbb{R}^{V\times C_v},
\)
and aggregating over all frames yields the module output $\mathbf{F}_j \in \mathbb{R}^{T\times V\times C_v}$. 

\subsection{Learned Soft Incidence Matrix Construction}
To represent and construct the hypergraph more flexibly than prior work, we propose to build a learned soft incidence matrix by pairing a bank of virtual hyperedge tokens $\mathbf{E}_v\in\mathbb{R}^{K_v\times C_v}$ with the temporally pooled joint summaries $\mathbf{F}_s\in\mathbb{R}^{V\times C_v}$. An illustrative example for $V=5$ and $K_v=5$ is shown in \cref{fig3b}.

Concretely, we compute the cosine similarity between each joint token and each virtual hyperedge token to obtain
\begin{equation}
\mathbf{S}\in\mathbb{R}^{V\times K_v},\qquad
S_{j,k} \;=\; \frac{\langle \mathbf{F}_s^{(j)},\;\mathbf{E}_v^{(k)}\rangle}
{\|\mathbf{F}_s^{(j)}\|\,\|\mathbf{E}_v^{(k)}\|},
\end{equation}
where $\mathbf{F}_s^{(j)}$ is the pooled feature of joint $j$ and $\mathbf{E}_v^{(k)}$ is the $k$-th virtual hyperedge token. We then apply a row-wise sparse normalisation to induce sparsity and produce the soft incidence matrix,
\(
\mathbf{H}_{\mathrm{soft}} \;=\; \operatorname{sparsemax}_{\text{row}}(\mathbf{S}) \in [0,1]^{V\times K_v},
\)
so that each row sums to one and entries are in $[0,1]$. The sparsity encourages compact joint-hyperedge memberships and reduces spurious associations.

In this formulation, a nonzero entry $\mathbf{H}_{\mathrm{soft}}(j,k)>0$ indicates that joint $j$ is assigned to hyperedge $\mathbf{E}_v^{(k)}$. Conversely, a column of $\mathbf{H}_{\mathrm{soft}}$ that is identically zero denotes an absent hyperedge, which is then ignored during subsequent hyperedge-to-joint feature redistribution.

\subsection{Single-Head Sample-Adaptive Transformer}

The SHSAT is a single-head transformer encoder \cite{vaswani2017attention} whose attention mask is generated per-sample from the soft incidence matrix $\mathbf{H}_{\mathrm{soft}}$. The mask enforces that joint features are aggregated onto their corresponding hyperedge tokens. Therefore, when a joint is not assigned to a given virtual hyperedge, it is masked out during attention. \cref{fig3b} shows a simple example of the mask construction. Since SHSAT uses a single layer and we retain only the first $K_v$ output tokens (the updated virtual hyperedge tokens) for downstream use, only the first $K_v$ rows of the mask are essential, and the remaining rows are inconsequential.

We also impose a visibility prior bias to suppress information flow from invisible joints. Let $\mathbf{vis}\in[0,1]^{T\times V}$ denote the joint visibility over an entire sequence, where $T$ is the number of frames and $V$ is the number of joints, with entries equal to $1$ for visible joints and $0$ for invisible ones. To limit contributions from invisible joints, we compute a per-entry bias by taking the natural logarithm after adding a small stability constant $\varepsilon>0$, \ie, \ $\mathbf{bias}=\log(\mathbf{vis}+\varepsilon)$. This bias is added to the attention logits before the softmax so that joints with low visibility receive negligible attention. It is worth noting that the entries of $\mathbf{vis}$ may take non-integer values. As the network depth increases, the MS-TCN progressively downsamples the temporal dimension, and the visibility map is downsampled accordingly to maintain alignment with the feature representation. This temporal downsampling may result in fractional visibility values.

In practice, the masked, visibility-biased single-head attention is computed as
\begin{equation}
\mathrm{Attention}(Q,K,V)
=\mathrm{softmax}\!\Big(\frac{QK^\top}{\sqrt{d}}+\mathbf{M}+\mathbf{B}_{\text{vis}}\Big)V,
\end{equation}
where $Q$, $K$, and $V$ are the query, key, and value matrices derived from linear projections of the input tokens, and $\mathbf{B}_{\text{vis}}$ contains the log-visibility terms broadcast to the corresponding joint-to-hyperedge logits.

\subsection{Training Strategy}

\subsubsection{Curriculum learning.}
We adopt curriculum learning \cite{bengio2009curriculum} to gradually expose the model to harder visibility conditions. Samples with most joints inside the FoV are treated as easy, while samples with most joints outside the FoV are treated as hard. Training starts with easy samples and progressively reduces the observable FoV to synthesise harder samples as training proceeds.

\subsubsection{Temporal CutMix.}
During training, we apply temporal CutMix \cite{yun2019cutmix} with probability $0.3$. Given two samples $a$ and $b$, we randomly select a contiguous segment of length $t$ from $a$ and a complementary segment of length $T-t$ from $b$, and concatenate them to form a mixed sample. 
The cross-entropy losses computed from the model logits on the mixed sample with respect to $a$'s label and $b$'s label are denoted as $\mathcal{L}_a$ and $\mathcal{L}_b$, respectively.
The final loss for the mixed sample is the length-weighted combination:
\begin{equation}
\mathcal{L}_{\text{mix}} \;=\; \frac{t}{T}\mathcal{L}_a \;+\; \frac{T-t}{T}\mathcal{L}_b.
\end{equation}

\subsection{Training Loss}

We train the model using the standard cross-entropy loss:
\begin{equation}
\mathcal{L}_{\mathrm{ce}}
=
-\sum_{c=1}^{N_{\text{class}}} y_c \log \hat{y}_c,
\end{equation}
where $\hat{y_c}$ denotes the predicted logits after softmax, $y_c$ is the one-hot ground-truth label, and $N_{\text{class}}$ is the number of action classes. 

In addition to $\mathcal{L}_{\mathrm{ce}}$, we design dedicated loss terms for the virtual hyperedge tokens $\mathbf{E}_v$ and the soft incidence matrix $\mathbf{H}_{\mathrm{soft}}$ to stabilise hypergraph construction and encourage meaningful hyperedge-to-joint assignments.

\subsubsection{Hyperedge Diversity Loss.}
Let the virtual-hyperedge bank be $\mathbf{E}_v\in\mathbb{R}^{K_v\times C_v}$ (denoted as a pool $P$ of $K_v$ vectors of dimension $C_v$). To encourage diversity among the $K_v$ vectors in the pool and avoid redundant hyperedges, we apply an orthogonality loss $ \mathcal{L}_{\text{pool}}$. 
For a single pool $P\in\mathbb{R}^{K_v\times C_v}$, first normalise each row to unit $\ell_2$ length (with a small constant $\varepsilon$ for stability) and form the Gram matrix
\begin{equation}
\tilde{P}_i=\frac{P_i}{\|P_i\|_2+\varepsilon},\qquad
\mathbf{G}=\tilde{P}\,\tilde{P}^\top\in\mathbb{R}^{K_v\times K_v}.
\end{equation}
The orthogonality loss for this pool is the mean squared off-diagonal entry of $\mathbf{G}$,
\begin{equation}
\mathcal{L}_{\text{pool}}(P)
\;=\;\frac{1}{K_v(K_v-1)}\sum_{i\neq j} G_{ij}^2
\;=\;\frac{\| \mathbf{G} - \mathbf{I}\|_{F}^2}{K_v(K_v-1)}.
\end{equation}

\subsubsection{Assignment Regularisation Loss.}
Let \( \mathbf{H}_{\mathrm{soft}}\in\mathbb{R}^{V\times K_v} \) denote the soft incidence matrix for one sample, where \(V\) is the number of joints and \(K_v\) is the number of virtual hyperedges. We introduce a composite assignment loss \( \mathcal{L}_{\text{assign}} \) that prevents degenerate and overly imbalanced assignments.

First, compute column sums over joints, 
\(
s_k = \sum_{j=1}^{V} (\mathbf{H}_{\mathrm{soft}})_{j,k} 
\), 
\(\mathbf{s}=(s_1,\dots,s_{K_v}),
\)
and normalise to obtain a distribution over edges
\begin{equation}
\mathbf{p} = \frac{\mathbf{s}}{\sum_{k=1}^{K_v} s_k + \varepsilon}.
\end{equation}

The loss comprises two terms:

1. \textbf{Batch-level balance.} Average \(\mathbf{p}\) over the batch to get \(\bar{\mathbf{p}}\), and penalise deviation from the uniform distribution \(\mathbf{u}=(1/K_v,\dots,1/K_v)\):
\begin{equation}
\mathcal{L}_{\text{balance}} \;=\; \frac{1}{K_v}\sum_{k=1}^{K_v} \big(\bar{p}_k - u_k\big)^2.
\end{equation}

2. \textbf{Per-sample max-hinge.} For each sample, take the maximum assignment probability \(\max_k p_k\) and apply a hinge loss with threshold \(\tau\) :
\begin{equation}
\mathcal{L}_{\text{hinge}} \;=\; \mathbb{E}\big[\max\big(0,\; \max_k p_k - \tau\big)\big],
\end{equation}
where the expectation is over samples.

The combined assignment loss is the weighted sum
\begin{equation}
\mathcal{L}_{\text{assign}}
\;=\; w_{\text{balance}}\mathcal{L}_{\text{balance}}
\;+\; w_{\text{hinge}}\mathcal{L}_{\text{hinge}},
\end{equation}
where \(w_{\text{balance}}\) and \(w_{\text{hinge}}\) denote the corresponding weighting coefficients. This regulariser encourages balanced and non-collapsed hyperedge assignments while allowing inactive (all-zero) columns to occur.

\subsubsection{Class-Centre Clustering Loss.}

We project the flattened soft incidence matrix of each sample into a compact embedding and denote the resulting per-sample features by \(\mathbf{z}_{n}\in\mathbb{R}^C\). The loss enforces that \(\mathbf{z}_{n}\) clusters around learnable class centres while keeping distinct class centres separated. Centres are maintained as parameters \(\mathbf{c}_{k}\in\mathbb{R}^C\) for each class \(k\).
The loss has two terms:
\begin{equation}
\mathcal{L}_{\text{pull}} \;=\; \frac{1}{N}\sum_{n=1}^{N}\big\| \mathbf{z}_{n} - \mathbf{t}_{n}\big\|_2^2,
\end{equation}
where \(N\) denotes the number of samples in the mini-batch. The target \(\mathbf{t}_{n}\) is the (possibly mixed) centre corresponding to the sample label:
\(\mathbf{t}_n=\lambda\,\mathbf{c}_{k_a}+(1-\lambda)\,\mathbf{c}_{k_b}\) to handle CutMix, where \(\lambda\) denotes the proportion of sample \(a\) in the mixed input (and \(\lambda=1\) when CutMix is not applied).

The repel term pushes different class centres apart. We compute pairwise centre distances and apply a hinge repulsion:
\begin{equation}
\mathcal{L}_{\text{repel}} \;=\; \frac{1}{\binom{N_{\text{class}}}{2}}\sum_{i<j}\mathrm{ReLU}\big(\Delta - \|\mathbf{c}_{i}-\mathbf{c}_{j}\|\big)^2.
\end{equation}
Here \(\Delta\) is the repel margin.

The combined loss is
\begin{equation}
\mathcal{L}_{\text{cluster}} \;=\; \mathcal{L}_{\text{pull}} \;+\; \gamma\,\mathcal{L}_{\text{repel}},
\end{equation}
with repel weight \(\gamma\). 

\section{Experiments}
\subsection{Datasets}

\subsubsection{NTU RGB+D.}
The NTU RGB+D dataset \cite{shahroudy2016ntu} contains 60 action categories, including daily activities such as drinking water, eating, brushing teeth, and throwing objects, with a total of 56,880 video samples. The data were collected from 40 subjects under 155 camera viewpoints using the Kinect v2 sensor, which simultaneously captured RGB, infrared, depth, and 3D skeleton sequences. Two standard evaluation protocols are provided, namely Cross-Subject (X-Sub 60) and Cross-View (X-View 60). Among the 60 action classes, only 11 correspond to two-person interaction actions, while the majority are single-person activities.

\subsubsection{NTU RGB+D 120.}
NTU RGB+D 120 \cite{liu2019ntu} is an extended version of NTU RGB+D, expanding the number of action categories to 120 and increasing the dataset size to 114,480 video samples. The data cover 106 subjects and 155 camera viewpoints. The introduced categories include more complex daily behaviours and sports-related activities, such as wearing headphones, shooting a basketball, and moving heavy objects. This dataset adopts Cross-Subject (X-Sub 120) and Cross-Setup (X-Set 120) evaluation protocols, placing greater emphasis on generalisation across different acquisition environments and device configurations. Among the 120 action classes, approximately 26 involve two-person interactions, while the remaining classes are predominantly single-person actions.

\subsection{Experimental Settings}
Due to the difficulty of collecting diverse human activity data~\cite{rajendran2024review}, we simulate constrained FoV scenarios by applying a FoV window to complete skeleton sequences.
The generation of constrained FoV scenarios involves two key factors: the selection of the FoV centre and the size of the FoV window.

To define the FoV centre, we select a joint from the first frame as the centre point. Since the choice of centre largely determines which body parts remain visible, we construct three types of test data by selecting representative joints, namely the head joint, the spine joint, and the base-of-spine joint, as the FoV centre. In addition, we create a fourth test setting where the centre joint is randomly sampled per sample, with detailed results provided in the supplementary material. 
The size of the FoV window directly determines how many joints remain visible. To systematically evaluate the model under varying visibility constraints, we design three levels of difficulty: easy, medium, and hard. A smaller window corresponds to a more challenging scenario. In the easy setting, approximately 75\% of the joints in a skeleton sequence are visible. In the medium and hard settings, only about 50\% and 25\% of the joints are visible, respectively.

\begin{table}[tb]
\caption{
Performance comparison on NTU RGB+D under constrained FoV settings with different FoV centres. Results are reported on easy, medium, and hard splits using X-Sub and X-View protocols. Bold font denotes the best result, and underline denotes the second-best result.
}
\label{tab1}
\centering
\renewcommand{\arraystretch}{1.0}
\setlength{\tabcolsep}{4pt}

\resizebox{0.9\linewidth}{!}{
\begin{tabular}{llcccccc}
\toprule
\multirow{2}{*}{FoV centre} & 
\multirow{2}{*}{Method} & 
\multicolumn{2}{c}{Easy (\%)} & 
\multicolumn{2}{c}{Medium (\%)} & 
\multicolumn{2}{c}{Hard (\%)} \\
\cmidrule(lr){3-4} \cmidrule(lr){5-6} \cmidrule(lr){7-8}
 &  & X-Sub & X-View & X-Sub & X-View & X-Sub & X-View \\
\midrule

\multirow{4}{*}{Head (3)}
& FR-Head \cite{zhou2023learning}  & 83.2 & 88.2 & 57.3 & 62.7 & 26.5 & 25.7 \\
& HD-GCN \cite{lee2023hierarchically}   & 81.9 & 85.9 & 53.4 & 58.7 & 23.5 & 24.7 \\
& Hyper-GCN \cite{zhou2025adaptive} & \underline{85.4} & \underline{91.3} & \underline{65.9} & \underline{72.4} & \underline{31.6} & \underline{34.4} \\
& Hyperformer \cite{zhou2022hypergraph} & 84.5 & 89.5 & 64.3 & 69.2 & 29.4 & 32.9 \\
& \textbf{PartialVisGraph} 
              & \textbf{90.9} & \textbf{95.8} 
              & \textbf{88.5} & \textbf{93.2} 
              & \textbf{79.2} & \textbf{84.5} \\
\midrule

\multirow{4}{*}{Spine (20)}
& FR-Head \cite{zhou2023learning}  & 80.6 & 85.3 & 51.3 & 54.7 & 21.7 & 20.5 \\
& HD-GCN \cite{lee2023hierarchically}   & 78.5 & 82.7 & 46.4 & 51.2 & 18.8 & 17.7 \\
& Hyper-GCN \cite{zhou2025adaptive} & \underline{82.1} & \underline{89.0} & \underline{58.5} & \underline{64.3} & \underline{24.1} & \underline{27.8} \\
& Hyperformer \cite{zhou2022hypergraph} & \underline{82.1} & 87.4 & 56.7 & 61.9 & 23.7 & 27.2 \\
& \textbf{PartialVisGraph} 
              & \textbf{90.6} & \textbf{95.4} 
              & \textbf{87.1} & \textbf{92.1} 
              & \textbf{77.7} & \textbf{83.2} \\
\midrule

\multirow{4}{*}{Base of the spine (0)}
& FR-Head \cite{zhou2023learning}  & 61.9 & 65.9 & 33.8 & 34.6 & 9.0 & 7.2 \\
& HD-GCN \cite{lee2023hierarchically}   & 57.3 & 59.1 & 30.8 & 29.9 & 7.2 & 6.6 \\
& Hyper-GCN \cite{zhou2025adaptive} & 65.1 & \underline{74.3} & 34.1 & 38.3 & 7.3 & 7.7 \\
& Hyperformer \cite{zhou2022hypergraph} & \underline{68.8} & 73.5 & \underline{40.0} & \underline{44.1} & \underline{10.8} & \underline{9.5} \\
& \textbf{PartialVisGraph} 
              & \textbf{89.2} & \textbf{94.5} 
              & \textbf{83.8} & \textbf{89.1} 
              & \textbf{71.4} & \textbf{76.5} \\
\bottomrule
\end{tabular}
}
\end{table}
\begin{table}[tb]
\caption{
Performance comparison on NTU RGB+D 120 under constrained FoV settings with different FoV centres. Results are reported on easy, medium, and hard splits using X-Sub and X-Set protocols. Bold font denotes the best result, and underline denotes the second-best result.
}
\label{tab2}
\centering
\renewcommand{\arraystretch}{1.0}
\setlength{\tabcolsep}{4pt}

\resizebox{0.9\linewidth}{!}{
\begin{tabular}{llcccccc}
\toprule
\multirow{2}{*}{FoV centre} & 
\multirow{2}{*}{Method} & 
\multicolumn{2}{c}{Easy (\%)} & 
\multicolumn{2}{c}{Medium (\%)} & 
\multicolumn{2}{c}{Hard (\%)} \\
\cmidrule(lr){3-4} \cmidrule(lr){5-6} \cmidrule(lr){7-8}
 &  & X-Sub & X-Set & X-Sub & X-Set & X-Sub & X-Set \\
\midrule

\multirow{4}{*}{Head (3)}
& FR-Head \cite{zhou2023learning}  & 74.2 & 75.5 & 45.9 & 46.0 & 12.7 & 14.5 \\
& HD-GCN \cite{lee2023hierarchically}   & 74.3 & 75.0 & 44.6 & 45.5 & 12.2 & 14.4 \\
& Hyper-GCN \cite{zhou2025adaptive} & 78.6 & \underline{81.8} & 52.4 & 54.2 & 12.7 & 16.8 \\
& Hyperformer \cite{zhou2022hypergraph} & \underline{79.0} & 80.3 & \underline{57.1} & \underline{55.1} & \underline{21.0} & \underline{19.8} \\
& \textbf{PartialVisGraph} 
              & \textbf{87.8} & \textbf{89.2} 
              & \textbf{84.1} & \textbf{85.9} 
              & \textbf{72.9} & \textbf{74.4} \\
\midrule

\multirow{4}{*}{Spine (20)}
& FR-Head \cite{zhou2023learning}  & 69.2 & 69.8 & 40.3 & 39.9 & 10.0 & 10.9 \\
& HD-GCN \cite{lee2023hierarchically}   & 68.5 & 68.9 & 38.7 & 38.3 & 9.2 & 10.9 \\
& Hyper-GCN \cite{zhou2025adaptive} & 74.4 & \underline{77.4} & 43.8 & 44.0 & 9.1 & 12.1 \\
& Hyperformer \cite{zhou2022hypergraph} & \underline{75.5} & 76.1 & \underline{50.5} & \underline{47.7} & \underline{16.6} & \underline{15.9} \\
& \textbf{PartialVisGraph} 
              & \textbf{87.2} & \textbf{88.6} 
              & \textbf{82.1} & \textbf{83.9} 
              & \textbf{71.5} & \textbf{73.1} \\
\midrule

\multirow{4}{*}{Base of the spine (0)}
& FR-Head \cite{zhou2023learning}  & 53.1 & 47.6 & 29.0 & 23.9 & 4.3 & \underline{3.9} \\
& HD-GCN \cite{lee2023hierarchically}   & 51.3 & 47.4 & 24.8 & 22.3 & 2.8 & 2.7 \\
& Hyper-GCN \cite{zhou2025adaptive} & 58.6 & 58.7 & 24.6 & 22.6 & 1.6 & 2.0 \\
& Hyperformer \cite{zhou2022hypergraph} & \underline{63.8} & \underline{60.5} & \underline{32.6} & \underline{30.0} & \underline{4.7} & 3.4 \\
& \textbf{PartialVisGraph} 
              & \textbf{85.3} & \textbf{86.7} 
              & \textbf{79.2} & \textbf{80.9} 
              & \textbf{66.7} & \textbf{67.4} \\
\bottomrule
\end{tabular}
}
\end{table}

\subsection{Implementation Details}

During training, we feed complete skeleton sequences to the model with a probability of 0.1, while randomly generated samples under constrained FoV are used with a probability of 0.9. As curriculum learning is adopted, the visibility ratio of these samples is gradually adjusted.
In the first epoch, the visibility ratio is set between 95\% and 100\%. After 70 epochs, it progressively decreases to a range of 20\% to 80\%, and remains unchanged thereafter. 
The model is trained for 200 epochs in total, with the first 5 epochs serving as a warm-up. The initial learning rate is set to 0.05 and is multiplied by 0.1 at the 110th and 120th epochs. We optimise the model using stochastic gradient descent (SGD) with Nesterov momentum of 0.9 and weight decay of 0.0004. The batch size is set to 64.

\subsection{Experimental Results}
\subsubsection{Experiments under Constrained Field-of-View.}

Multi-stream ensemble methods are widely adopted to boost performance in prior works \cite{chi2022infogcn,zhou2024blockgcn,shi2019skeleton}, and in our experiments, the ensemble remains beneficial under constrained FoV, \cref{tab1} and \cref{tab2} report results using the commonly used 4-stream ensemble. Results show that, regardless of which joint is selected as the FoV centre and regardless of the severity of the constrained FoV, PartialVisGraph consistently performs well. In particular, under the hard difficulty level, almost all competing methods collapse, whereas PartialVisGraph maintains stable performance.

\subsubsection{Experiments under Full Field-of-View.}
\begin{table}[tb]
  \caption{
  Performance comparison on NTU RGB+D and NTU RGB+D 120 under full FoV settings.
  For fair comparison, all methods adopt a 4-stream ensemble.
  Bold font denotes the best result, and underline denotes the second-best result.
  }
  \label{tab3}
  \centering
  \setlength{\tabcolsep}{4pt}
  \renewcommand{\arraystretch}{1.0}

  \resizebox{\textwidth}{!}{
  \begin{tabular}{l l cc cc cc}
    \toprule
    \multirow{2}{*}{Method} & 
    \multirow{2}{*}{Publication} &
    \multirow{2}{*}{Params (M)} &
    \multirow{2}{*}{GFLOPs} &
    \multicolumn{2}{c}{NTU RGB+D (\%)} & 
    \multicolumn{2}{c}{NTU RGB+D 120 (\%)} \\
    \cmidrule(lr){5-6} \cmidrule(lr){7-8}
     &  &  &  & X-Sub & X-View & X-Sub & X-Set \\
    \midrule
    CTR-GCN \cite{chen2021channel}     & ICCV 2021  & 1.5 & 1.97 & 92.4 & 96.4 & 88.9 & 90.6 \\
    Hyperformer \cite{zhou2022hypergraph} & arXiv 2022 & 2.6 & -- & 92.9 & 96.5 & 89.9 & 91.3 \\
    FR-Head \cite{zhou2023learning}    & CVPR 2023  & 2.0 & -- & 92.8 & 96.8 & 89.5 & 90.9 \\
    HD-GCN \cite{lee2023hierarchically}     & ICCV 2023  & 1.7 & 1.77 & 93.0 & 97.0 & 89.8 & 91.2 \\
    SkateFormer \cite{do2024skateformer} & ECCV 2024  & 2.0 & 3.62 & 93.5 & \textbf{97.8} & 89.8 & 91.4 \\
    BlockGCN \cite{zhou2024blockgcn}   & CVPR 2024  & 1.3 & 1.63 & 93.1 & 97.0 & 90.3 & 91.5 \\
    ProtoGCN \cite{liu2025revealing}   & CVPR 2025  & -- & -- & 93.5 & 97.5 & 90.4 & 91.9 \\
    Hyper-GCN \cite{zhou2025adaptive}  & ICCV 2025  & 2.3 & 2.88 & \underline{93.7} & \textbf{97.8} & \underline{90.9} & \underline{92.0} \\
    \midrule
    \textbf{PartialVisGraph} & -- 
    & 4.0
    & 2.90 
    & \textbf{93.8} 
    & \underline{97.6} 
    & \textbf{91.0} 
    & \textbf{92.3} \\
    \bottomrule
  \end{tabular}
  }
\end{table}

We train and evaluate PartialVisGraph on full FoV skeleton sequences, and for fair comparison, we adopt the widely used 4-stream ensemble strategy. \cref{tab3} presents a comprehensive comparison on NTU RGB+D and NTU RGB+D 120 against a large set of benchmark methods. PartialVisGraph outperforms all compared approaches, demonstrating its effectiveness and superiority even with complete skeleton observations, and showing that our learnable hypergraph construction and feature aggregation remain advantageous on full skeleton inputs.

\subsection{Ablation Study}

\begin{table}[tb]
  \caption{Ablation results under full FoV and constrained FoV settings}
  \label{tab4}
  \centering
  \setlength{\tabcolsep}{5pt}
  \renewcommand{\arraystretch}{1.0}
  \footnotesize
  \begin{tabular}{p{4.8cm} c p{3.5cm} c}
    \toprule
    \multicolumn{2}{c}{Full FoV (\%)} &
    \multicolumn{2}{c}{Constrained FoV (\%)} \\
    \cmidrule(lr){1-2} \cmidrule(lr){3-4}
    
    PartialVisGraph & 92.0 &
    PartialVisGraph & 78.8 \\

    Wo $\mathcal{L}_{\text{pool}}$ & 91.8 &
    Wo $\mathbf{B}_{\text{vis}}$ & 78.4 \\

    \parbox[t]{3.8cm}{
    Wo $\mathcal{L}_{\text{assign}}$ and $\mathcal{L}_{\text{cluster}}$
    } & 91.2 &
    Wo Curriculum learning & 78.4 \\

    Wo CutMix & 91.0 &
    -- & -- \\
    
    \bottomrule
  \end{tabular}
\end{table}
\begin{figure}[tb]
  \centering
  \begin{subfigure}{0.23\linewidth}
    \centering
    \includegraphics[width=\linewidth]{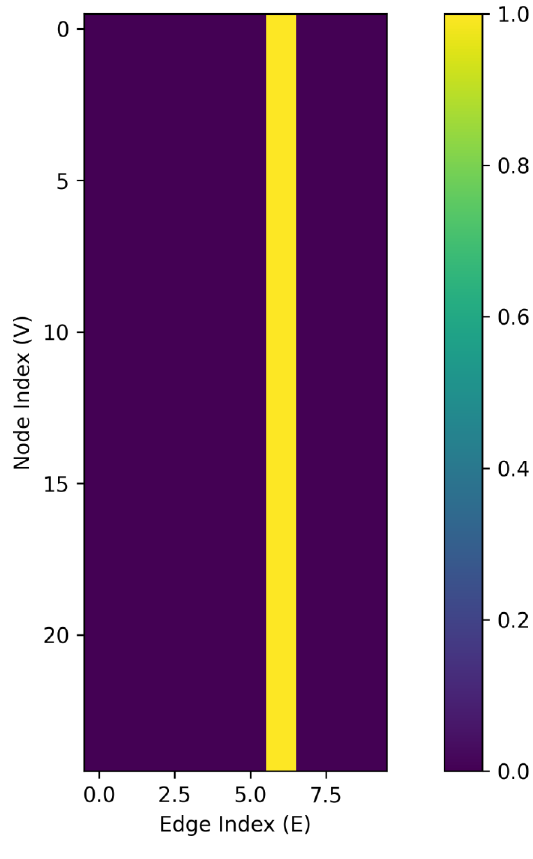}
    \caption{Soft incidence matrix without $\mathcal{L}_{\text{assign}}$}
    \label{fig4a}
  \end{subfigure}
  \hfill
  \begin{subfigure}{0.23\linewidth}
    \centering
    \includegraphics[width=\linewidth]{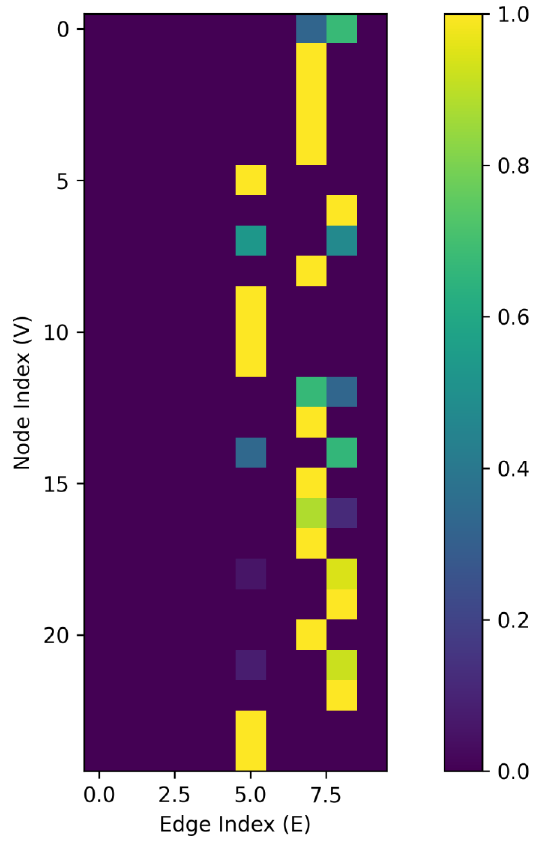}
    \caption{Soft incidence matrix with $\mathcal{L}_{\text{assign}}$}
    \label{fig4b}
  \end{subfigure}
  \hfill
  \begin{subfigure}{0.5\linewidth}
    \centering
    \includegraphics[width=\linewidth]{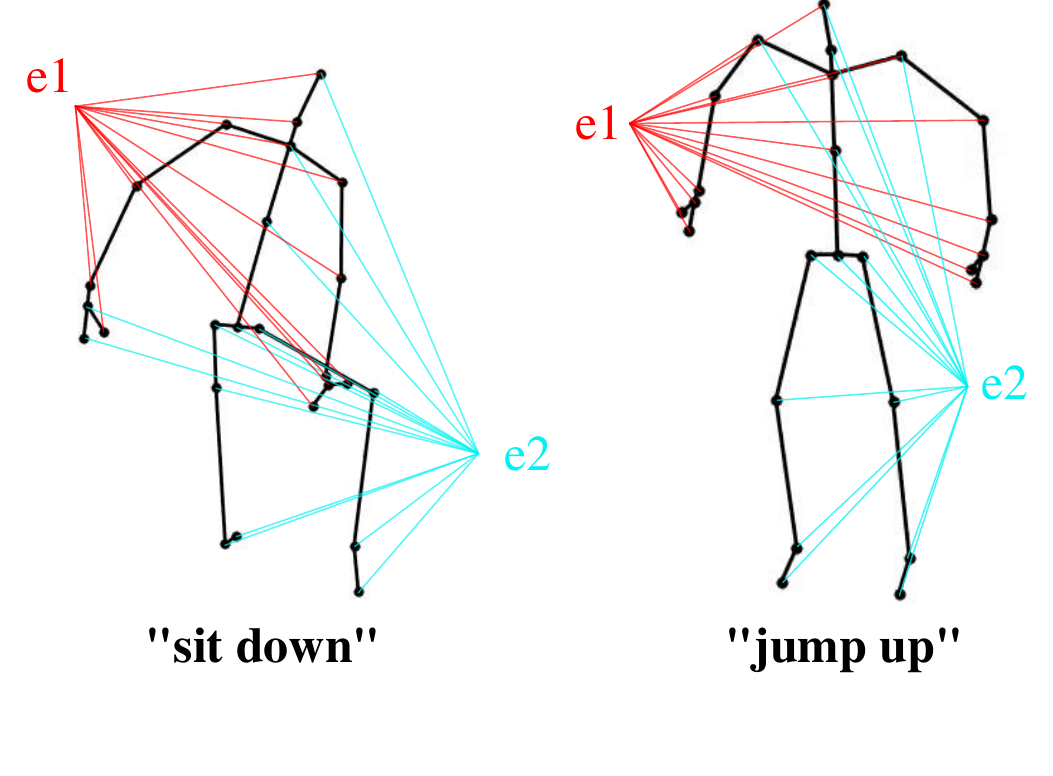}
    \caption{Constructed hypergraph of our model}
    \label{fig4c}
  \end{subfigure}
  \caption{Visualisation of the hypergraph}
  \label{fig:abstudy}
\end{figure}

All ablation experiments are evaluated on the joint modality. Since temporal CutMix, $\mathcal{L}_{\text{pool}}$, $\mathcal{L}_{\text{assign}}$ and $\mathcal{L}_{\text{cluster}}$ apply to both constrained and full FoV data, \cref{tab4} reports their effects under full FoV. The constrained FoV results shown are averaged across the three difficulty levels.

Removing $\mathcal{L}{\text{pool}}$ leads to a 0.2\% drop in accuracy, indicating that promoting diversity among hyperedges contributes to performance gains. When $\mathcal{L}{\text{assign}}$ and $\mathcal{L}_{\text{cluster}}$ are removed jointly, accuracy decreases by 0.8\%. As shown in \cref{fig4a}, the model then collapses to a single hyperedge covering all joints, confirming that these regularisation terms prevent assignment collapse. 
Further insight is provided in \cref{fig4c}, which visualises hypergraphs constructed in intermediate layers. Here, $e1$ and $e2$ denote two hyperedges connected to their member joints. 
The hypergraph construction realises a structural reorganisation of the skeleton. As illustrated by the two samples, they group joints into semantically coherent parts (one group comprising the legs and spine and another comprising both arms). This reorganisation provides a structured basis for subsequent feature aggregation across related joints.

From \cref{tab4}, we further observe that omitting temporal CutMix reduces accuracy by $1\%$, suggesting that temporal CutMix contributes positively to the model's performance. 
Under constrained FoV, disabling the visibility prior lowers accuracy by $0.4\%$, highlighting the importance of the visibility prior. Finally, curriculum learning yields a $0.4\%$ accuracy gain, indicating that progressively introducing more challenging visibility conditions benefits the model's performance on the test set.

\section{Conclusion}

In this paper, we investigated skeleton-based human action recognition under constrained FoV, and proposed a novel hypergraph-based framework that learns soft incidence matrices via a bank of virtual hyperedge tokens, aggregates joint features onto hyperedges with a Single-Head Sample-Adaptive Transformer (SHSAT), and applies a visibility prior to restrict information flow from invisible joints. 
To ensure a thorough evaluation, we design multiple experimental protocols that mimic varying degrees of view restriction and benchmark our method against recent state-of-the-art approaches. Empirical results show that PartialVisGraph achieves leading performance across these protocols. Moreover, when trained and evaluated under the full FoV setting, it also outperforms all compared methods.

\section*{Acknowledgements}
This work was supported in part by the National Natural Science Foundation of China (62576152, 62332008), the Basic Research Program of Jiangsu (BK20250104), and the Fundamental Research Funds for the Central Universities (JUSRP202504007).

%
%
\bibliographystyle{splncs04}
\bibliography{main}
\end{document}